\documentclass{article}

\AtBeginDocument{%
  }
    
 \usepackage[final, nonatbib]{neurips_2024}

\usepackage[utf8]{inputenc} 
\usepackage[T1]{fontenc}    
\usepackage{hyperref}       
\usepackage{url}            
\usepackage{booktabs}       
\usepackage{amsfonts}       
\usepackage{nicefrac}       
\usepackage{microtype}      
\usepackage{xcolor}         
\usepackage{wrapfig,lipsum,booktabs}
\usepackage{svg}
\usepackage{graphicx}

\usepackage{biblatex}
\title{Contextual Evaluation of Large Language Models for Classifying Tropical and Infectious Diseases}

\author{%
Mercy Asiedu\thanks{corresponding author: masiedu@google.com} \\
 Google Research \\
  \And
  Nenad Tomasev  \\
    Google DeepMind \\
\\
    \AND
  Chintan Ghate  \\
  Google Research \\
    \And
   Tiya Tiyasirichokchai\\
    Google Research \\
     \\
    \And
    Awa Dieng \\
    Google DeepMind \\
     \And
    Oluwatosin Akande \\
    Independent Scientist \\
    \And
    Geoffrey Siwo \\
    University of Michigan \\
    \And
    Steve Adudans \\
    Gear Health \\
    \And
    Sylvanus Aitkins \\
    Ministry of Health, Sierra Leone\\
    \And
    Odianosen Ehiakhamen \\
    Nigerian Center for Disease Control \\
    \And
    Eric Ndombi \\
    Kenyatta University \\
    \And
    Katherine Heller \\
    Google Research \\
}

\begin{document}

\maketitle

\begin{abstract}
  While large language models (LLMs) have shown promise for medical question answering, there is limited work focused on tropical and infectious disease-specific exploration. We build on an opensource tropical and infectious diseases (TRINDs) dataset, expanding it to include demographic and semantic clinical and consumer augmentations yielding 11000+ prompts. We evaluate LLM performance on these, comparing generalist and medical LLMs, as well as LLM outcomes to human experts. We demonstrate through systematic experimentation, the benefit of contextual information such as demographics, location, gender, risk factors for optimal LLM response. Finally we develop a prototype of TRINDs-LM, a research tool that provides a playground to navigate how context impacts LLM outputs for health. 
\end{abstract}

\section{Introduction}
\label{sec:intro}
 Neglected tropical diseases, while extremely preventable and treatable, continue to be highly prevalent in the poorest regions of the world, affecting 1.7 billion people globally with disproportionate impacts on women and children \cite{hudu2024insight}. Similarly, half of the world's population are at risk from infectious diseases, which continue to lead in global mortality and morbidity, resulting in more than 52 million deaths globally, with 99\% of diseases occurring in developing countries \cite{rana2023changes}.  Challenges in preventing and treating these diseases include surveillance, early detection, accurate initial diagnosis, management and vaccine limitations \cite{hudu2024insight, vuitika2022vaccines}. During the pandemic various scalable measures were implemented to address these challenges specific to COVID-19 \cite{bragazzi2020big, gupta2020analysis, mbunge2020integrating, marbouh2020blockchain, lee2020global}.  Other tropical and infectious diseases have had limited attention for surveillance and surfacing accurate diagnosis.

 The use of large language models for health-related question-answering has increased in recent years demonstrating applications for a variety of health use cases \cite{sallam2023chatgpt, tian2024opportunities, lu2024large, mannhardt2024impact}. However, there is limited work that has focused on tropical and infectious diseases. These are particularly of interest as they may present out-of-distribution cases, given that they mostly occur in the global south which may be underrepresented in training and evaluation datasets and may lead to potential biases \cite{asiedu2023globalizing, asiedu2024case}. Additionally there is limited understanding of how different contextual factors such as demographics, prompt styles, and subsets of information (eg. symptoms only, versus symptoms+location) may influence model performance. 

A few studies assess the use of machine learning and LLMs for tropical and infectious diseases. \textcite{shenoy2022artificial} found that in a study of 40 tropical and infectious disease clinicians, 35 indicated the need for a decision making tool. They compared logistic regression and decision trees for binary classification and found an average prediction accuracy of 79-84\%. \textcite{pfohl2024toolbox} developed a health equity dataset which introduced LLM evaluation of a tropical and infectious diseases (TRINDs) dataset and had clinicians and health equity experts evaluate LLM responses to tropical and infectious disease related questions,finding evidence of biased responses.  \textcite{mondal2024evaluating} developed 50 simulated infectious disease cases with histories, lab reports, imaging findings and evaluated it on 4 different general LLMs and compared the quantity of differential diagnosis to that of medical experts. They found that LLMs generally had difficulties matching the experts’ differential diagnosis. \textcite{schwartz2024black} delineate failure points for LLMs for infectious diseases consultation to clinical workflow questions and find that LLMs provide dangerous hallucinations and harmful advice for disease management-type questions.

In this paper we build on the TRopical and INfectious Diseases (TRINDs) dataset \cite{pfohl2024toolbox}, making the following primary contributions:

\begin{itemize}
    \item We expand on the dataset to include demographic and semantic clinical and consumer augmentations.
    \item We perform various evaluations with the dataset, to understand how different contexts and counterfactual locations contribute to LLM performance.
    \item We evaluate LLM performance improvements on the larger augmented dataset with in-context prompt tuning.
    \item We assemble a panel of human experts to set a human expert baseline score on the dataset and to provide ratings of data quality, usefulness, etc. 
    \item We develop TRINDs-LM, a research tool to demonstrate how context influences LLM performance on TRINDs. 
\end{itemize}

\section{Methods}
\begin{table*}[hbtp]
  {\caption{Summary of datasets and experiments}}
  \label{tab: methods-summary}
  \centering
  {\begin{tabular}{ll}
  \toprule
  \bfseries Dataset augmentations & \bfseries Experiments \\
  \midrule
 Original TRINDs dataset & Generalist LLM vs. specialist LLM accuracy\\  &  LLM vs. human expert performance
\\
Contextual dataset & Impact of contextual factors on accuracy\\
French dataset & Impact of language on accuracy\\
Counterfactual dataset & Impact of location, race and gender on accuracy\\
Multiple choice set & LLM vs. human expert performance\\
LLM-expanded demographic set & Impact of a variation of demographics on accuracy\\
LLM-expanded semantic set & Impact of a variation of question semantic styles on accuracy\\
Consumer set & Impact of consumer style questions on accuracy\\
  \bottomrule
  \end{tabular}}
\end{table*}

\subsection{Dataset generation and expansion:}
\textit{\textbf{Original TRINDs dataset:}} We base this work on the original TRINDs dataset \cite{pfohl2024toolbox}. To summarize, the authors examined authoritative sources containing factual information about different diseases to create a dataset of 106 questions pertaining to tropical and infectious diseases across different regions around the world. We use a subset of 52 questions from the TRINDs data as a seed set. Each question in the seed set follows a templated structure where a patient persona is presented with general symptoms, direct attributes, specific symptoms, as well as context and lifestyle/risk factors. Each question is associated with a ground truth disease label, which were reviewed by clinicians for accuracy. 

\textit{\textbf{Contextual TRINDs dataset:}}We created 16 subsets of the seed set to understand which of the different sections of the dataset most influenced model performance. This set used different inclusions and combinations of general symptoms, specific symptoms, personal attributes (age and gender), location, and risk factors.This generated 468 queries.

\textit{\textbf{Counterfactual sets:}} \textit{Location:} We examined how location influences LLM responses. The original dataset was created with locations where the disease had a known likelihood of occurring. For each of the original dataset and contextual subsets that included a location, we switched out the original location for a single counterfactual location where there was less probability of disease occurring. Here we used “San Francisco” as that location to generate 52 counterfactual location queries. \textit{Race:} We created versions of the original dataset where we included a race input (Asian, Black and White) for each disease, yielding 159 additional queries. \textit{Gender:} We created versions of the dataset that had only male, only female and non-binary demographics, yielding an additional 159 queries. 

\textit{\textbf{French language set:}} Given prevalence of tropical and infectious diseases in non-English speaking countries we sought to understand how language influenced the performance of the model. A researcher whose official language is French, manually translated the original dataset of 52 queries and diagnosis into French and compared performance to the English dataset. We selected French as our primary interest lies in  the African continent and official languages that are used are primarily in English and French. This led to 52 prompts. 

\textit{\textbf{Multiple choice set:}} To compare the LLMs to expert baselines, we generate 153 multiple choice questions from the original, expanded demographics and  expanded semantics sets. We input the questions and ground truth diagnosis from the original dataset and prompted an LLM to provide 5 multiple choice options that included the ground truth label, and 4 other broad tropical and infectious disease options. 

\textit{\textbf{LLM-Expanded demographics set:}}
To provide a larger demographically diverse data pool to assess LLM responses we created a synthetically expanded dataset of 2635 queries using the original set as a seed set. For each persona an LLM was prompted to generate 50 demographic expansions, with varying gender, sex, age, socioeconomic status, disability status, ethnicity, location, and origin. For the location variation the LLM was instructed to only generate locations with high incidence of the disease. The LLM was also instructed to prioritize including locations that were consistent with the generated demographics, and to adjust the socioeconomic status to match the location that was generated. Finally, each proposed generation was checked by an LLM-based filter where the LLM was asked to discard generations where the symptoms do not match the condition, or where the demographic does not appear to be plausible. Refer to prompting method in Appendix \ref{apd:third}.

\textit{\textbf{LLM-Expanded semantic set:}} As prompts from users may present in a variety of ways, and as symptoms may vary as well, we synthetically expanded the dataset, adding 2651 additional queries. These were created by instructing the LLM to first generate a demographically altered version of the question, in the same way as described above for the expanded demographics set. The LLM then follows that up by another alteration, generating an alternative formulation of the question for the given patient and disease, but describing a different set of common symptoms related to the disease, and potentially a different personal patient background of relevance for disease risk or onset. LLM-based filtering was applied to these generations as well, where proposals would get discarded in case the generated symptoms did not match the disease, or the proposed demographic does not appear to be plausible.

\textit{\textbf{LLM-Consumer augmented queries:}} The original personas were created with a clinical tone. We created a consumer augmentation of the expanded dataset. We prompted an LLM to convert the original, expanded demographics and semantic sets into first person perspective to generate the consumer versions. This generated 52 consumer queries on the original set, 2635 on the demographic and 2651 queries on the semantic set.

In total we generated a dataset of 11,719 queries from the original seed set of 52 queries across 50 tropical and infectious diseases. We summarize the dataset and experiments run in Table \ref{tab: methods-summary}. Refer to Appendix \ref{apd:second} for examples.

\subsection{Model evaluation}
\label{sec:model_eval}
We use two baseline models- Gemini Ultra  \cite{gemini}, a generalist large language model (standard hyperparameters: batchsize 16, temperature 0.7, top\_k=32), and MedLM Medium \cite{medlm}, a LLM specialized for the health domain (standard hyperparameters: batchsize 32, temperature 0.2, top\_k=None). As a baseline, we prompt-tuned both models, providing instructions and 2-shot examples  to guide the model’s output (Appendix \ref{apd:third}).  The models were prompted to provide an output of the ground truth disease label.  This was repeated 4 additional times to yield a total of 5 outcomes per experiment for statistical analysis.

\subsection{Auto-rater LLM  Evaluations}
We developed an automated rater to score each query for accuracy out of the 5 repeated runs. This was developed by prompting an LLM to determine whether the words were structurally and/or meaningfully similar to each other and to score them as correct if they were. For instance if the ground truth was \textit{Taeniasis/cysticercosis} and model output was \textit{Tapeworm} it would be marked as correct, since these are meaningfully similar, even if they are structurally different. We performed a manual review of the automated rating on a subset of the data to optimize the rating process and ensure there were no errors.

From the scores we first determined performance of generalist and medical specific models on variations of the original persona set. Next we analyzed which contextual factors/combinations were important for model performance (combinations of general symptoms, specific symptoms, risk factors, location and personal attributes). We then compare performance on the counterfactual and original versions of the prompt. Finally, we assess performance on expanded data with demographic and semantic augmentations with and without many-shot prompting with the original set. 

\subsection{Human Expert Baseline}
\label{sec:human_Experts}

\begin{wraptable}{r}{5.5cm}
  \caption{Expert panel demographics (n=7).   \label{tab:expert}}
  \centering
  {\begin{tabular}{ll}
  \toprule
  \bfseries Demographic & \bfseries No. Experts \\
  \midrule
 \textbf{Gender} &  \\
     Female  & 2 \\
     Male & 5
    \\
    \textbf{Age} & \\
    30-39 & 2\\
    40-49 & 5\\
    \textbf{Country of residence} & \\
    Kenya & 2\\
    Sierra Leone & 1\\
    United States & 2\\
    Nigeria & 2 \\
    Switzerland & 2 \\
    \textbf{Highest level of education} & \\
    Masters & 2\\
    MD & 2\\
    Doctorate & 3\\
    \textbf{Occupation} & \\
    Medical doctor & 4\\
    Public health researcher & 4\\
    Professor & 1\\
    \textbf{Years of experience} & \\
    5-10 & 3\\
    11-20 & 3\\
    20+ & 1\\
  \bottomrule
  \end{tabular}}
  \vspace{-4mm}
\end{wraptable}

We created a human expert baseline study to understand expert performance on the dataset. The purpose of this was to determine how experts in tropical and infectious diseases performed on a representative sample of the data to enable human expert vs LLM comparison. The study was deemed IRB exempt by an internal ethics review personnel. We recruited experts- public health researchers, and medical doctors- who had generalizable knowledge of TRINDs,  using a snowballing approach via our networks.

Following informed consent, the experts filled a demographic pre-survey, summarized in Table \ref{tab:expert}. Experts reported a variety of specializations with tropical and infectious diseases including  Immunoparasitology, Neglected Tropical Diseases, Infectious Disease Epidemiology, and diseases with pandemic/epidemic potential (eg. Ebola, Mpox).

Experts were then given 52 short answer questions (SAQs) with full context- general and specific symptoms, demographics (age and gender), location, and risk factors across the identified diseases- and asked to write in a single most likely diagnosis. Once they completed the SAQs, they were given another questionnaire with multiple choice questions of varied formats (153)  where they selected the single most likely disease given a list of diseases. Experts were asked not to reference any sources in answering the questions or to look up answers to the questions until both surveys had been completed. After completing the SAQ and MCQ surveys, experts provided feedback on various axes on data quality. They were also asked to indicate how helpful each contextual information was in answering the questions. The process took ~5 hours all together and experts were compensated 500 USD for their time.

\subsection{TRINDs-LM Tool design and development}
We developed the TRINDs-LM playground for researchers to understand how context impacts health responses for LLMs, using Gemini Ultra as a base model.  The user interface (Appendix \ref{apd:first}) allows the user to input demographic information, lifestyle information, and symptoms. A summary is then generated and input into the LLM. The model provides an output that includes the most likely diagnosis, reason, and disease definition. The user interface also shows an interactive  map of the global disease incidence rate. The TRINDs research tool is available on request. The tool not meant for clinical use. 

\subsection{Statistical analysis}
We performed two-tailed student's t-test (alpha=0.025) for statistical analysis to compare performance between medical specific and general purpose models for each experiment, original and counterfactual datasets, base and many-shot prompt tuned models, and LLM and human expert performance.

\section{Results}

\begin{figure}[htbp]
    \centering
     \includegraphics[width=1.1\linewidth]{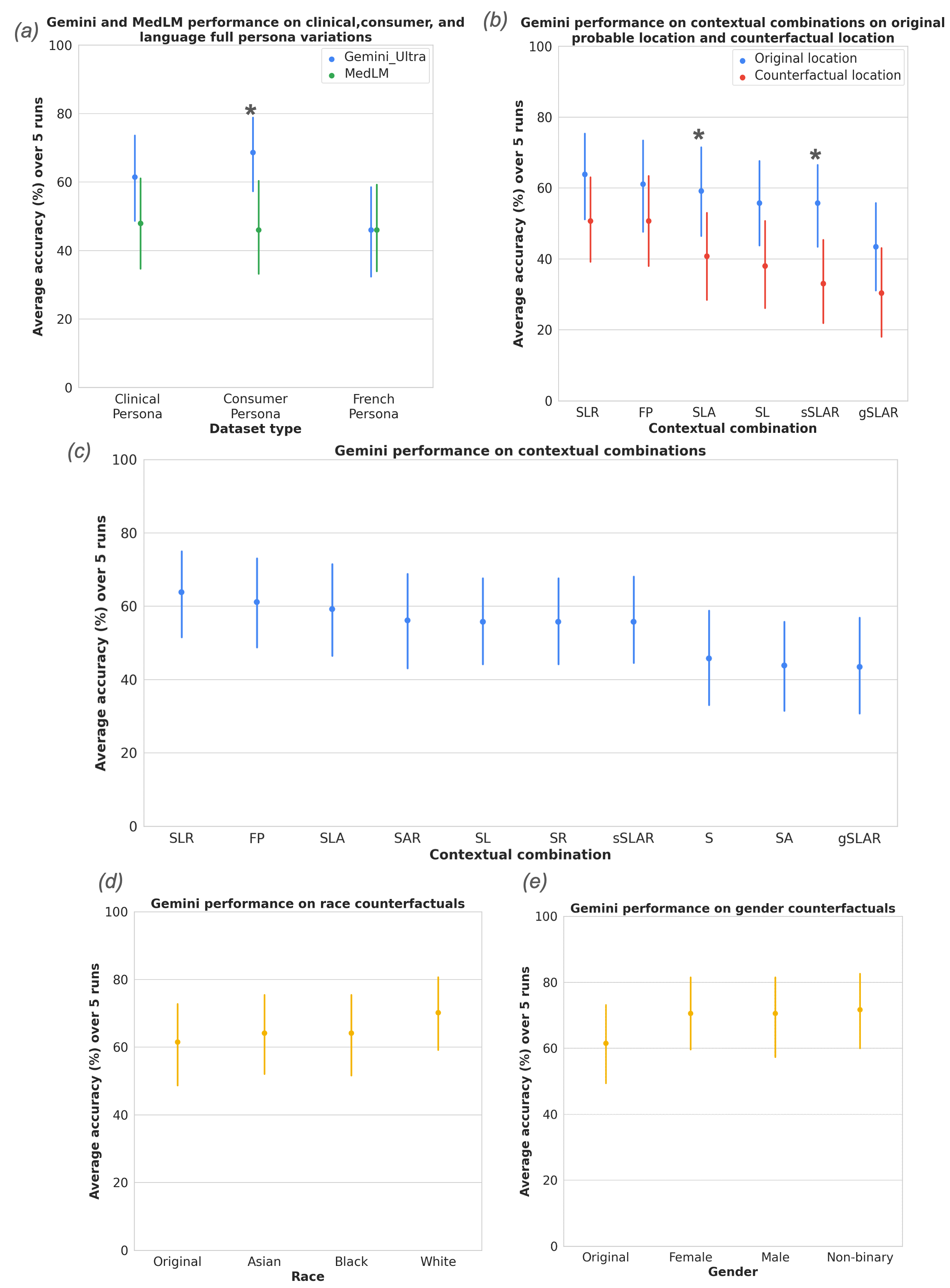}
    \caption{Model performance on persona variations. a) Generalist (Gemini) and specialist (MedLM) model performance on clinical, consumer and French persona variations. b) Gemini model performance on counterfactual location inputs. c) Gemini performance for contextual combinations of attributes and factors and count. d)   Gemini performance for race counterfactuals  e)  Gemini performance for gender counterfactuals.   
    \textbf{Legend:} S=symptoms(general and specific), gS=general symptoms, sS= specific symptoms, L=location, A=attribute (age and gender)
  Error bars are 90\%  confidence interval. *=p\textless0.025
}
     \label{fig:results1}

\end{figure}

\subsection{LLM experimental results}
\subsubsection{Generalist (Gemini Ultra) and specialist (MedLM Medium) model performance on persona variations}
We found that overall performance with minimum instruction tuning and two-shot examples yielded an accuracy of 61.5\% for Gemini Ultra and 47.9\% for MedLM for the original clinical personas (Figure \ref{fig:results1}a). Gemini Ultra performance increased to 68.7\% for the consumer versions, while MedLM remained roughly the same at 46.0\%. Performance of both models was reduced with the French versions of the questions,  with Gemini dropping significantly to 46.0\% compared to MedLM which dropped slightly to 46.0\%. Overall the generalist gemini model outperformed MedLM, which might be due to factors such as differences in model sizes, or overfitting of the MedLM tuned model to specific datasets (Figure \ref{fig:results1}a).

\subsubsection{Assessing Gemini model performance on varied combinations of attributes and factors}
\begin{figure}[htbp]
\centering
\includegraphics[width=0.9\linewidth]{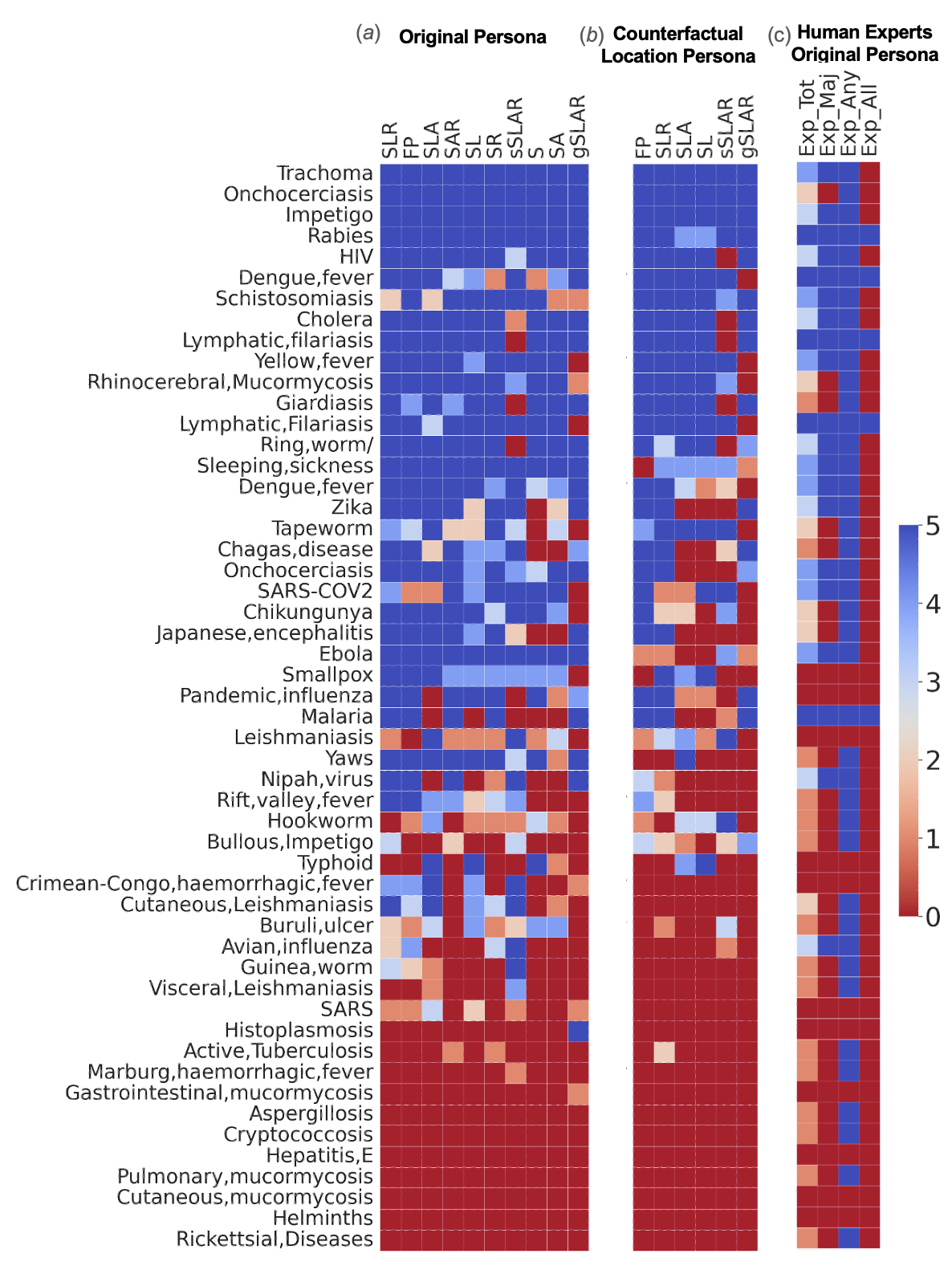}
    \caption{Per disease performance for LLMs and human experts, a) LLM performance on original persona with different contextual combinations (5 repeated runs), b) LLM performance  on location counterfactual with different contextual combinations (5 repeated runs), c) Human expert performance (top 5 out of 7 experts). 
  Error bars are 90\%  confidence interval. \textbf{Legend:} S=symptoms(general and specific), gS=general symptoms, sS= specific symptoms, L=location, A=attribute (age and gender), R=risk factor, FP=full persona, Exp\_Tot=total expert score, Exp\_Maj = expert majority score, Exp\_Any = expert any/at least one score, Exp\_All = Expert all score.  }      
     \label{fig:perdisease}

\end{figure}

We found that \textit{symptoms}, \textit{location} and \textit{risk factors (SLR)} enable the best model performance,  followed by the \textit{full persona (FP)} of symptoms, location, risk factors and personal attributes  (Figure \ref{fig:results1}c). This implies that excluding personal attributes such as age and gender, may preserve privacy while still maintaining beneficence. The worst performing contextual combinations were \textit{general and specific symptoms (S)} alone (46.8\%), \textit{symptoms} and \textit{attribute (SA)} (44.91\%) and \textit{general symptoms}, \textit{location}, \textit{attributes} and \textit{risk factors (gSLAR)} (44.5\%), demonstrating the need for both specific and general symptoms, as well as other contextual factors such as location and risk factors to enhance model accuracy.  We also show performance per disease and find that the model performs best on relatively widespread or distinctive diseases eg. Trachoma, HIV and Rabies, but performs worst on less common or diseases with common symptoms eg. tuberculosis (Figure \ref{fig:perdisease}a). 

\subsubsection{Assessing Gemini model performance on counterfactual inputs}
\textit{\textbf{Location:}}  We found that generally, the \textit{counterfactual location} decreased performance across all contextual combinations that included \textit{location (L)}, but performance was least affected in cases with \textit{full persona (FP)}, or with the combination of \textit{general and specific symptoms, location and risk (SLR)} demonstrating the usefulness of including added contextual information  (Figure \ref{fig:results1}b). \textit{Counterfactual location} caused the model to perform the worst when only \textit{symptoms} were provided without any additional context, or when either \textit{general symptoms} or \textit{specific symptoms} were provided even with other information. This demonstrates the need to have both \textit{specific} and \textit{general symptoms}, plus other contextual information for optimal performance. We find that this pattern is consistent across the individual diseases (Figure \ref{fig:perdisease}b). \textit{\textbf{Race:}} There were no statistically significant differences in performance across different racial counterfactual inputs, even though we note that performance for “White race” counterfactuals increased slightly (Figure \ref{fig:results1}d). \textit{\textbf{Gender:}} There were no statistically significant differences in performance across different gender  counterfactual inputs, even though we note that performance for each of the gender counterfactuals was slightly higher than the original persona (Figure \ref{fig:results1}e).

\subsubsection{Assessing model performance on the demographic and semantic expansions}
 We found no significant difference for base Gemini (Figure \ref{fig:tuned} a,b) performance for clinical and consumer variations on the expanded dataset. We found no significant differences between performance of the base models on demographic and semantic variations (Figure \ref{fig:tuned}a,b). This demonstrates that models perform approximately the same on clinical style and consumer style questions which contain the same information. 

\subsubsection{Impact of many-shot in-context learning with original persona set on model robustness and generalizability}
\begin{wrapfigure}{r}{0.5\textwidth}
\vspace{-25mm}
\centering
\includegraphics[width=0.9\linewidth]{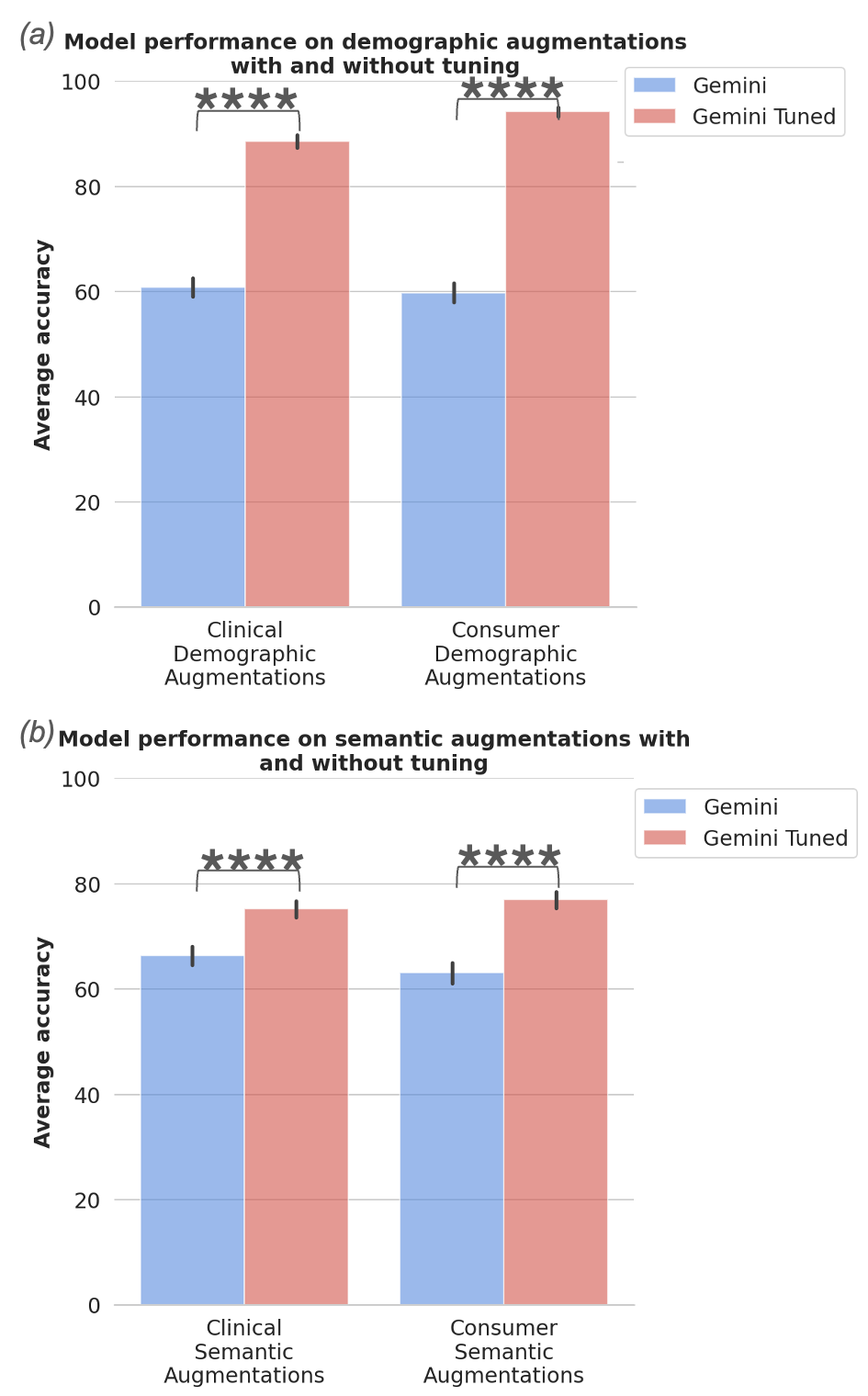}
    \caption{Model performance on expanded dataset. a) LLM performance on demographic clinical and consumer augmentations (2635 each) and b)LLM performance on semantic clinical and consumer augmentations (2651 each). We compared the base model, with the multi-shot tuned model. ****=p\textless0.00005}      
     \label{fig:tuned}
    \vspace{-5mm}
\end{wrapfigure}

We found that in-context learning by providing the model with many-shot examples of the full original set for each disease, significantly increased Gemini performance for demographic and semantic augmentations, though less so for semantic augmentations(Figure \ref{fig:tuned} a,b). It also increased performance on different styles of question inputs (semantics). This demonstrates that in-context learning with a small set of high quality data improves model performance and robustness across different demographics, locations and question styles.

\subsection{Human expert baseline and data rating}
\subsubsection{Expert vs LLM performance:}
Seven experts filled out the baseline SAQs and MCQs. We calculated the score for each expert and selected responses for the top 5 scoring experts for each survey for further analysis.  For each question from the top 5 responses, we calculated the (i) the \textit{Expert total score}: sum of the score from the 5 experts, (ii) \textit{Expert majority score}: full score (5/5) if the majority of experts had the correct answer else 0/5, (iii) \textit{Expert any/at least one score}: a full score (5/5) if any one of the experts had the correct answer, and (iv) \textit{Expert all score}: a full score (5/5) only  if all the experts had the answer correct. TRINDs experts tend to specialize and different specialists may do better on some diseases than others. These combinations  of expert scores simulate real-world policy settings where a panel of experts may be used to come to a diagnosis in a variety of ways.

For the SAQs dataset, Gemini performed significantly better than all the expert response combinations, except for the \textit{Expert Any/at least one} which was significantly better than Gemini. For the MCQs, Gemini significantly performed better than the \textit{Expert total} and \textit{Expert all scores}, and significantly worse than the \textit{Expert Any score}. There was no difference between Gemini and \textit{Expert majority} performance for MCQs (Figure \ref{fig:experts}a).

\subsubsection{Expert data rating and qualitative feedback:}
\begin{figure}[h]
\centering
  \includegraphics[width=1.1\linewidth]{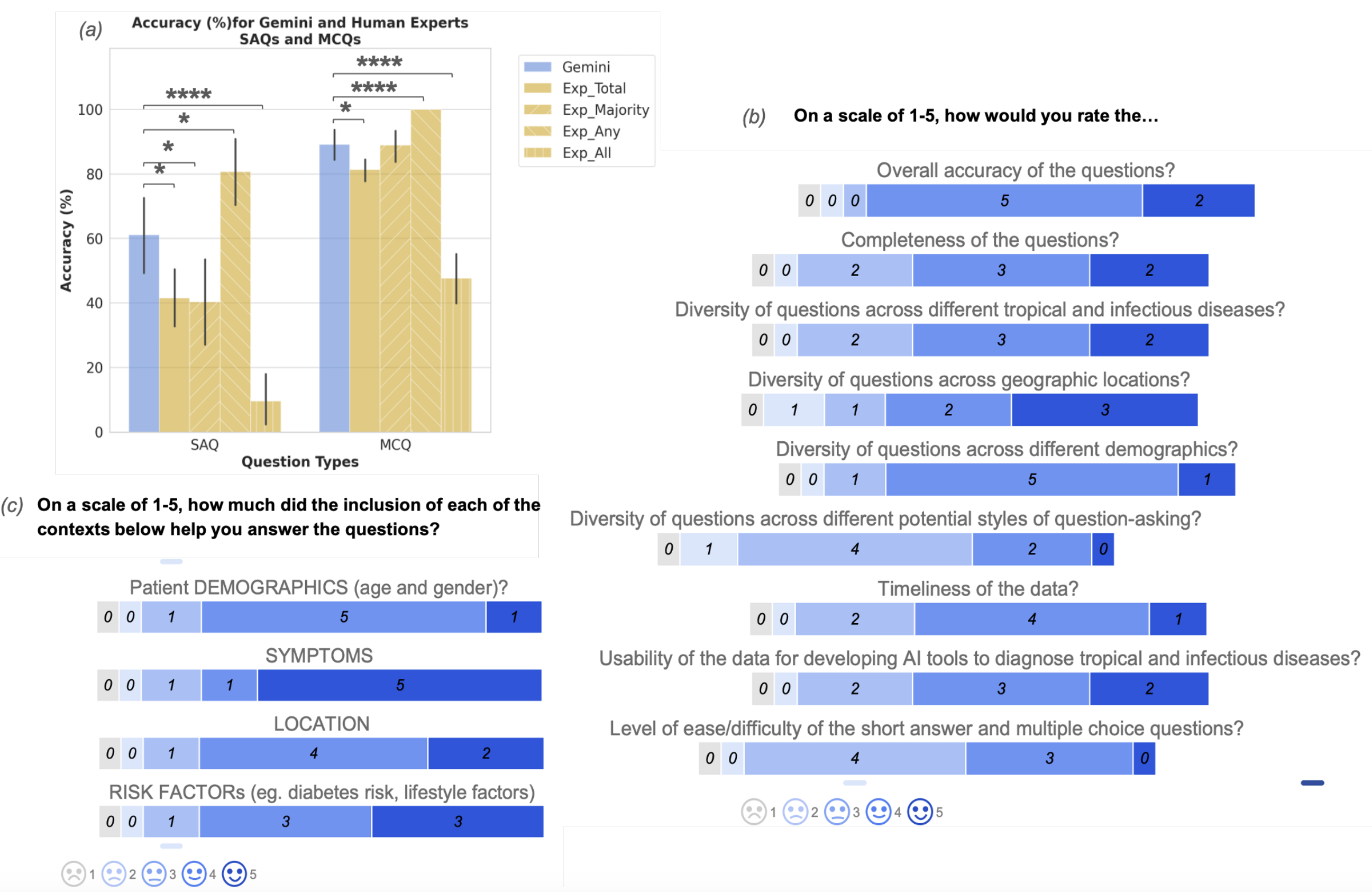}
  \caption{ Expert baseline and data quality rating. a)Expert baseline compared to LLM, b) Expert rating of data quality, c) Expert rating of helpfulness of contextual information.  
  Error bars are 90\% CI. *=p\textless0.05 ****=p\textless0.00005}
  \label{fig:experts}

\end{figure}

Experts rated level of helpfulness of each of the contextual information in providing a diagnosis. \textit{Symptoms} were rated highest by most experts, followed by \textit{risk factors}, \textit{location} and then demographics \textit{attributes} (Figure \ref{fig:experts} c). This is in line with LLM results that indicate that \textit{symptoms, location and risk factors} are most important for  LLM classification. On the dataset quality, experts generally indicated high scores (4-5) for accuracy, completeness, range of tropical and infectious diseases covered, geographic and demographic diversity, and timelines of the dataset (Figure \ref{fig:experts} b). Experts rated the level of difficulty of the questions as 3-4/5. The diversity in the style of questions-asking was rated low with several experts commenting on the repetitiveness of the questions. Experts recommended including the use of over-the-counter medications, less repetition in the question style, improving the specificity of some of the symptoms, and the need for differential diagnosis. There were also comments on improving the quality of the LLM-generated queries. One expert commented on the inclusion of smallpox which has been eradicated for 40 years, but indicated the potential tie to the ongoing Monkeypox (Mpox) epidemic. Experts also indicated the need for patient images to provide more informative responses demonstrating the need for a multimodal version of the dataset. 

\section{Discussion}
This work is motivated by the need to evaluate LLMs on tropical and infectious diseases which present a potential out-of-distribution shift for LLM performance given that these tend to occur in the global south which may have less representation in datasets used for LLM training \cite{asiedu2023globalizing, asiedu2024case}. This study confirms this distribution shift with the Gemini model achieving an accuracy of 61.5\% and MedLM achieving an accuracy of 47.9\% on clinical-style questions, significantly lower than reported performances on USMLE benchmarks (GPT: 90.2\%, MedLM: 91.1\% \cite{saab2024capabilities}). We find that simple in-context prompting with the dataset, and potentially intentional training focus on these diseases improves the LLM performance significantly. We find that Gemini Ultra performs better than MedLM medium, however this is likely due to differences in model sizes. 

Our evaluations demonstrate that including additional context such as risk factors and location in addition to symptoms also improves model performance. However indicating locations where the disease is less likely to occur, reduces model performance. This has implications for LLMs that may attempt to provide a disease diagnosis based on autorecognition of location without considering contexts such as visits to endemic locations. On this dataset, we do not find significant differences in LLM performance across race or gender. We also do not find any significant differences in performance between different styles of question asking (semantic, clinical, and consumer variations). 

 Our analysis reveal that LLMs tend to more accurately identify common diseases, or diseases with very specified symptoms such as \textit{Trachoma, Rabies} and \textit{Yaws}, while less common and less specific diseases are mislabeled. For instance we find that LLMs mostly classify \textit{Hepatitis E} as \textit{Hepatitis A}, though this was a common mistake among human experts as well.
 
 Our human expert baseline finds that for both SAQs and MCQs, experts scored lower in accuracy on the full context questions than the model except in cases where we looked at \textit{expert any/at least one score}. \textit{Avian influenza} is an example of a disease that the LLM had trouble with, but that majority of the experts got right. On the other hand the model got diseases like \textit{Giardiasis} and \textit{Onchocerciasis} right, while most experts got these wrong. We do note that experts were asked not to consult any external material, which would not be the case in real-world scenarios, and real world scenarios may also provide confirmatory tests for the diseases.
 
 Experts generally rated the dataset highly on axes of accuracy, completeness, timeliness and diversity across tropical and infectious diseases. However they suggested improvements in diversity in question asking styles, and addition of images to the questions where applicable.
 
 \textbf{Implications for Policy and Practice:} Our findings demonstrate the discrepancy in LLM performance on tropical and infectious diseases, compared to reported performance on USMLE questions, identifying the need for contextual evaluation of LLMs that are used in clinical settings in the global south - i.e the need for LLM usage to take into account contextual, regional-specific and patient-specific factors. However, we also find that Gemini performs better on this dataset compared to human experts for identifying TRINDs from text-based descriptions. For healthcare workers, our findings highlight the potential of LLMs to serve as valuable decision-support tools, thus enhancing clinical diagnostic accuracy in resource-limited settings. Notably, these tools should complement, not replace, clinical judgment and should be balanced by continuous evaluation and refinement of these models to maintain their relevance and reliability in diverse clinical settings.

\textbf{Limitations and Future work:}
\label{sec:limitations}
Limitations of this work include the focus on only disease classification, primary focus on English and primary focus on text-based queries. Future work could explore evaluating other tasks such as management steps and treatments, additional languages and multimodal datasets such as relevant disease-related images, or sounds from coughs and breathing. Another limitation is that we used a relatively small sample size of experts, which may not represent the breath of experience in this very broad subject area. Future work should look at a larger sample size of experts reflecting geographic and sub-specialty diversity to improve expert assessment and provide a more generalizable human expert baseline. 

\section{Conclusion}
 Overall this study finds that while LLM performance on providing a diagnosis of the tropical and infectious diseases dataset is low, we find that experts performance is similarly low in most cases. LLM performance improves with simple in-context learning with our dataset. We find that larger generalist models outperform smaller specialist models.  We underline the importance of context in performance, noting that providing symptoms, risk factors and location outperform the provision of symptoms alone to the LLM. This work provides a scalable methodology for evaluating LLMs for health in global settings for out-of-distribution cases.

\printbibliography

\appendix

\section{Appendix / supplemental material}

\subsection{Images of the TRINDs-LM interface}
\label{apd:first}

 \begin{figure}[htbp]
\centering
\includegraphics[width=1\linewidth]{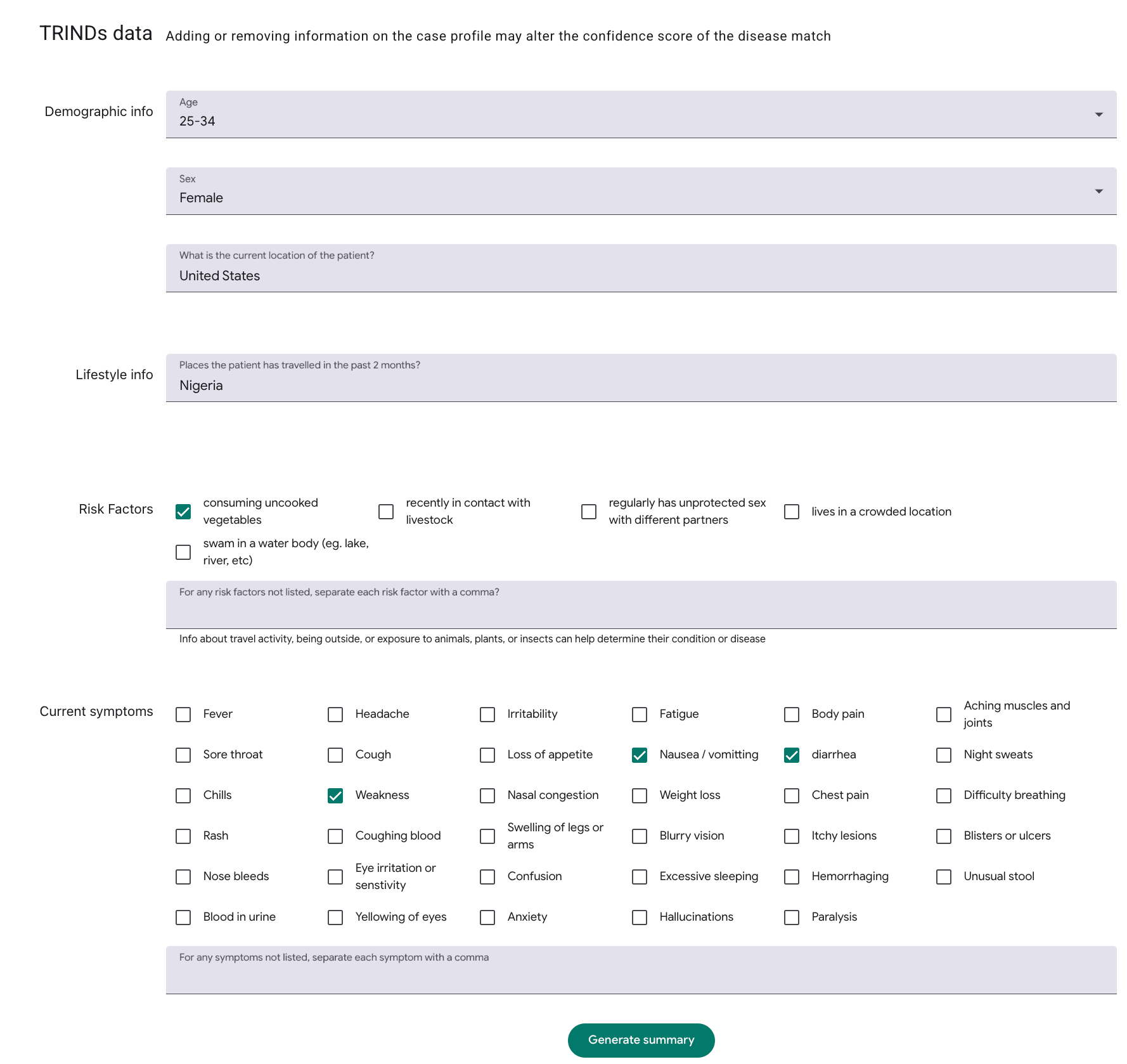}
  \label{fig:demo1}
  \caption{TRINDs research tool showing user entry}
\end{figure}

 \begin{figure}[htbp]
\centering
\includegraphics[width=1\linewidth]{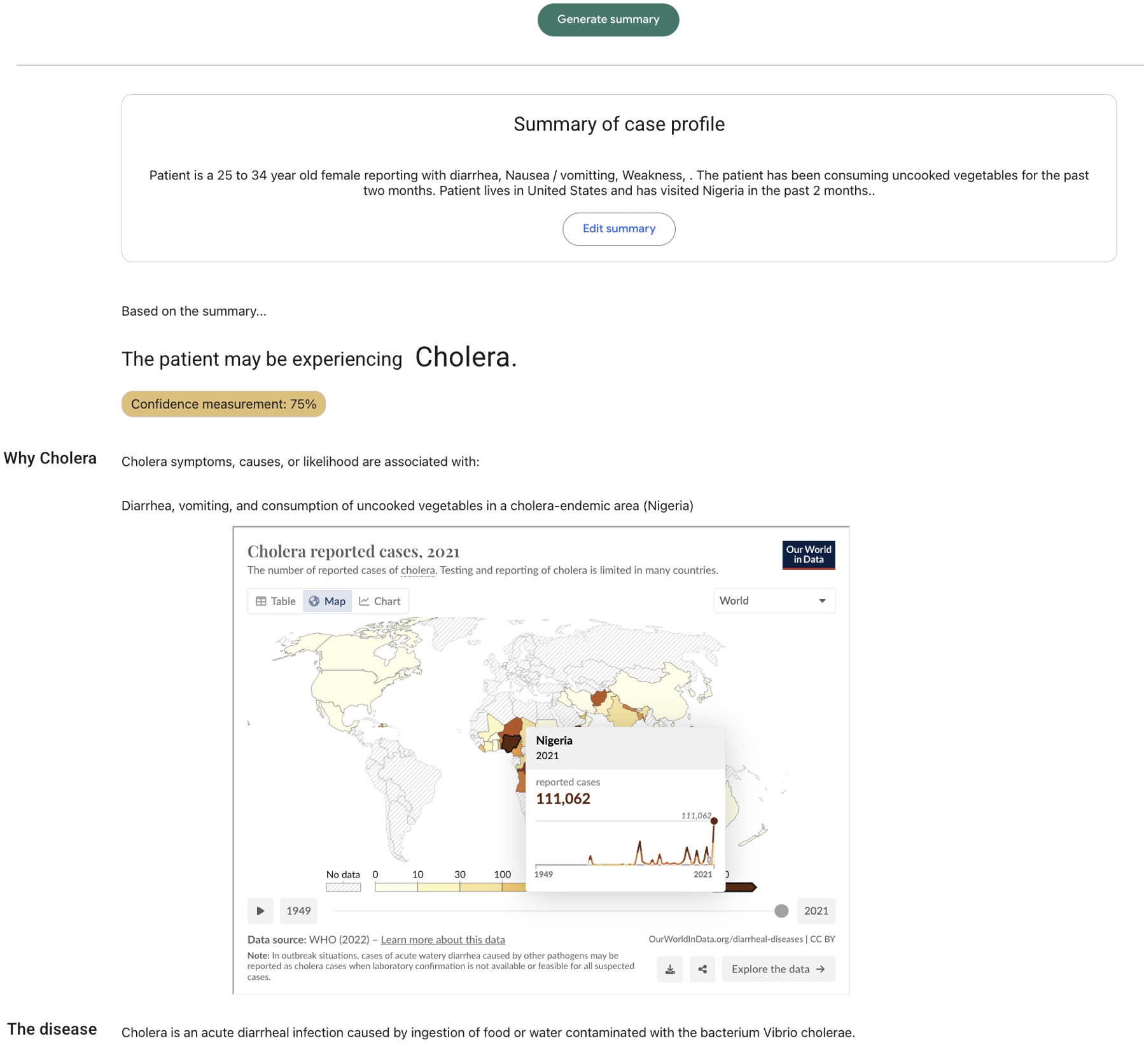}
  \label{fig:demo2}
  \caption{TRINDs research tool showing LLM output}
\end{figure}

 \clearpage
 
\subsection{Examples of data/query types}\label{apd:second}
This section shows the different data/query augmentations developed used for evaluations. 

\begin{figure}[htbp]
\centering
\includegraphics[width=1\linewidth]{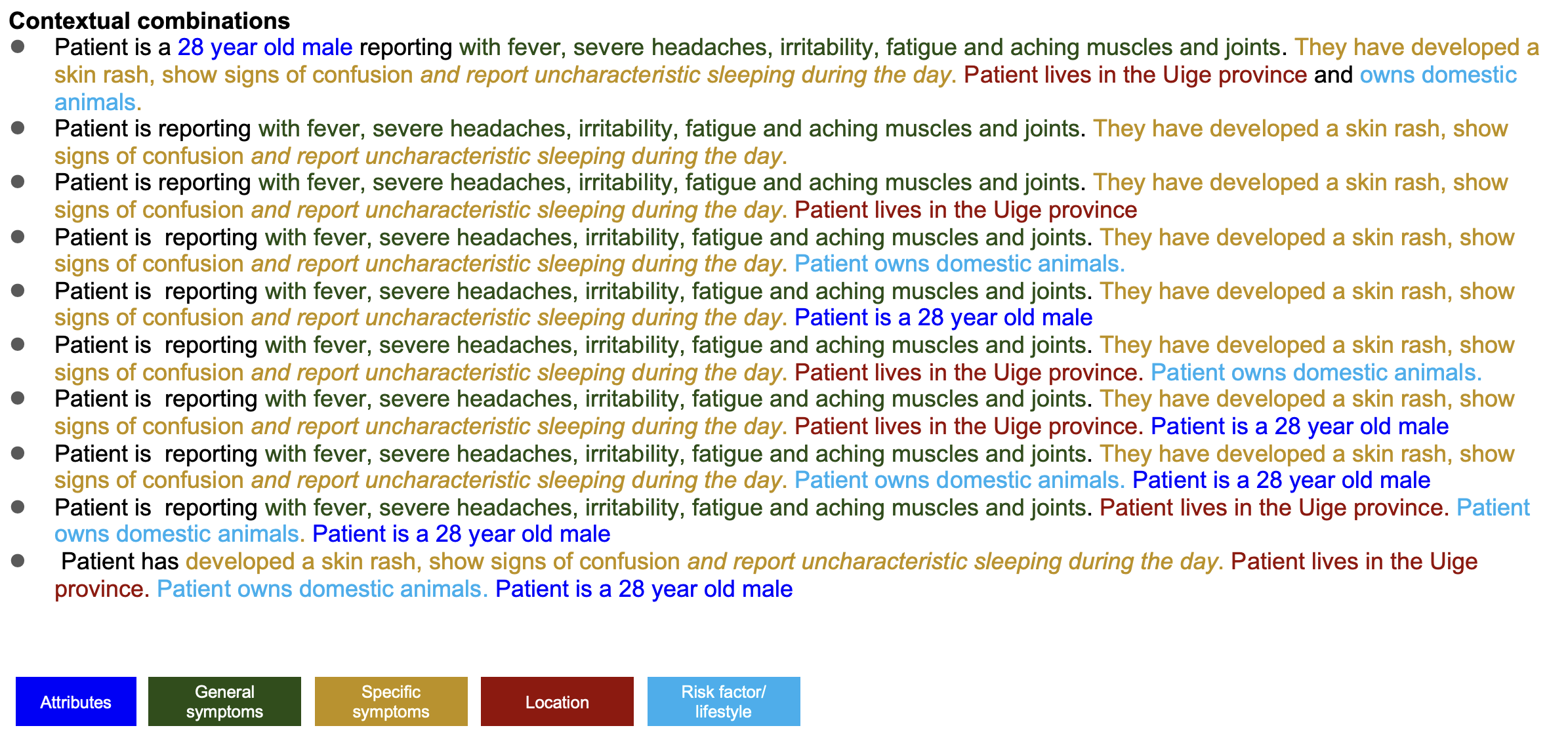}
  \label{fig:context_aug}
  \caption{Contextual combinations of general symptoms,specific symptoms, location, risk factor and attributes. The first combination is the original full persona containing all context}
\end{figure}

\begin{figure}[htbp]
\centering
\includegraphics[width=1\linewidth]{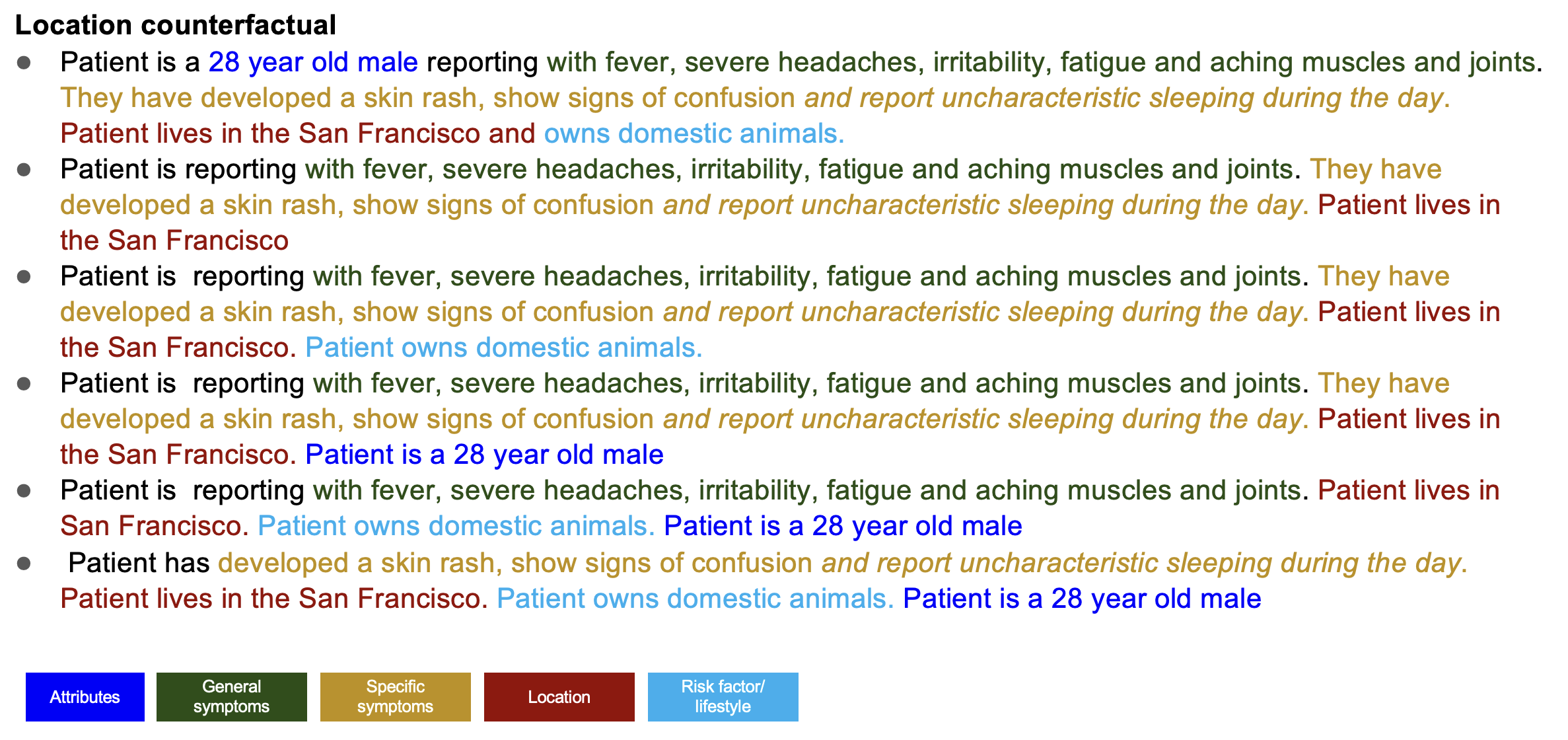}
  \label{fig:loc_cnt}
  \caption{Location counterfactual with context combination}
\end{figure}

\begin{figure}[htbp]
\centering
\includegraphics[width=1\linewidth]{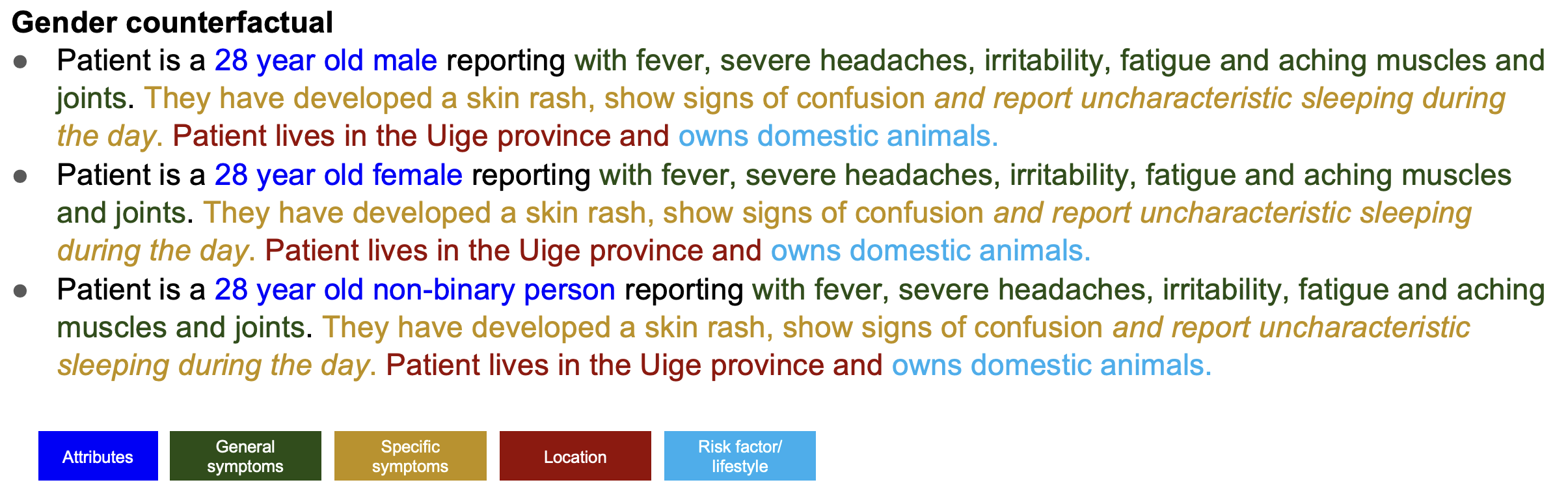}
  \label{fig:gender_cnt}
  \caption{Gender counterfactual of full persona}
\end{figure}

\begin{figure}[htbp]
\centering
\includegraphics[width=1\linewidth]{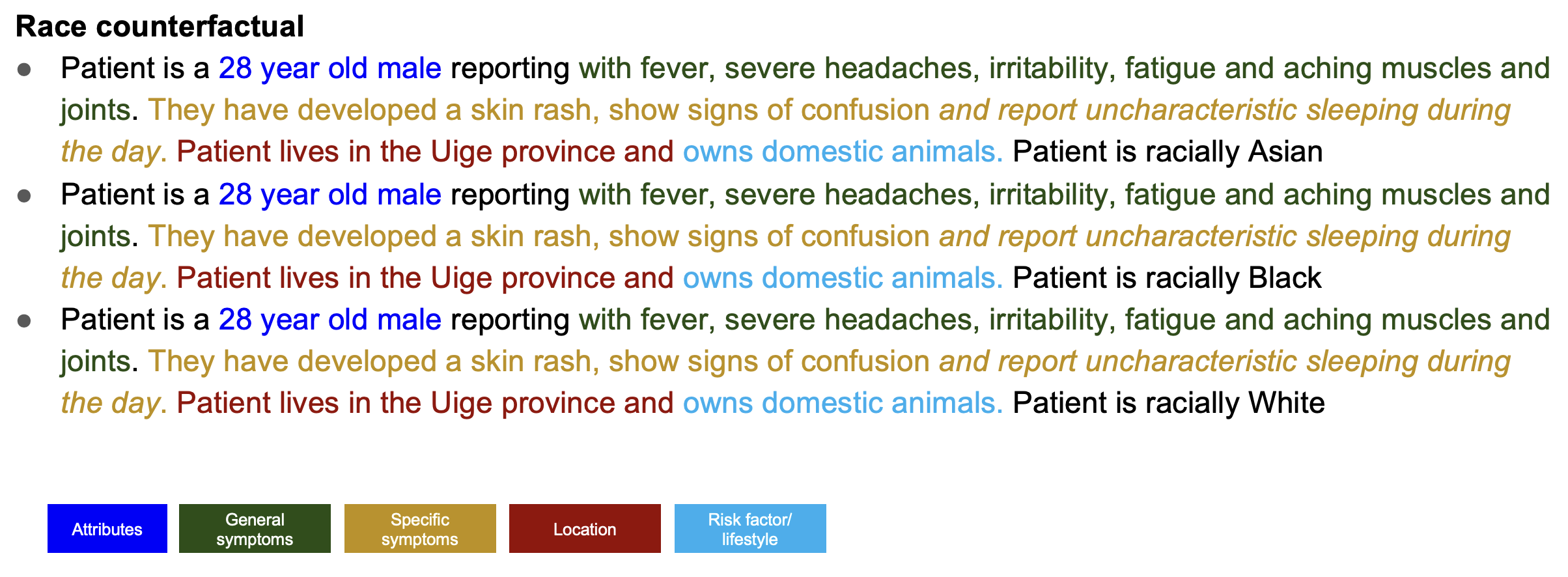}
  \label{fig:race_cnt}
  \caption{Race counterfactual of full persona}
\end{figure}

\begin{figure}[htbp]
\centering
\includegraphics[width=1\linewidth]{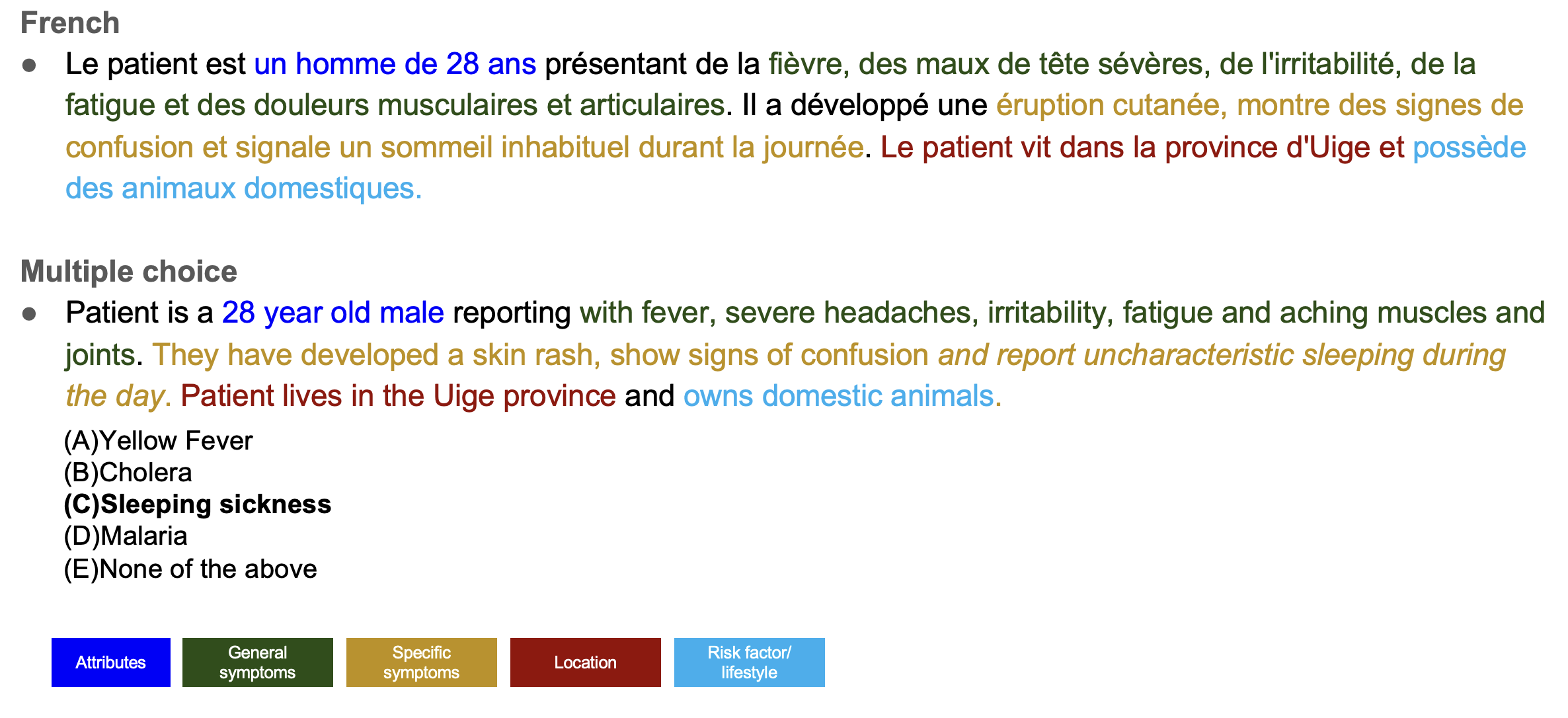}
  \label{fig:lang_cnt}
  \caption{Language (French) and Multichoice formats}
\end{figure}

\begin{figure}[htbp]
\centering
\includegraphics[width=1\linewidth]{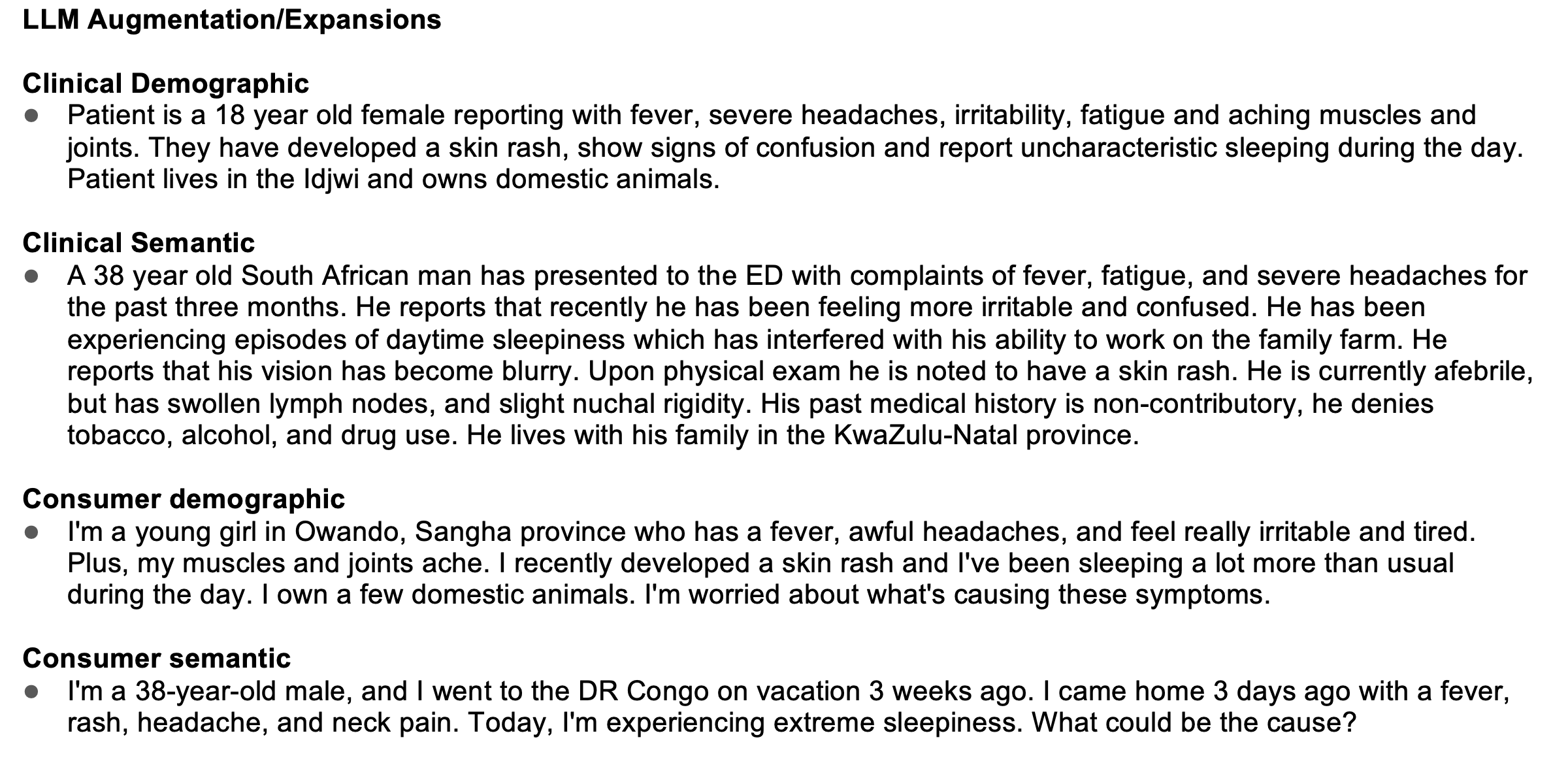}
  \label{fig:llm_aug}
  \caption{Language model augmentations/expansions}
\end{figure}
\clearpage

\subsection{Sample Prompts}
\label{apd:third}
This section shows prompts used to generate LLM classification for open ended and multiple choice questions. The section also shows sample prompts used to create the LLM demographic Augmentations/Expansions

\begin{figure}[htbp]
\centering
\includegraphics[width=1\linewidth]{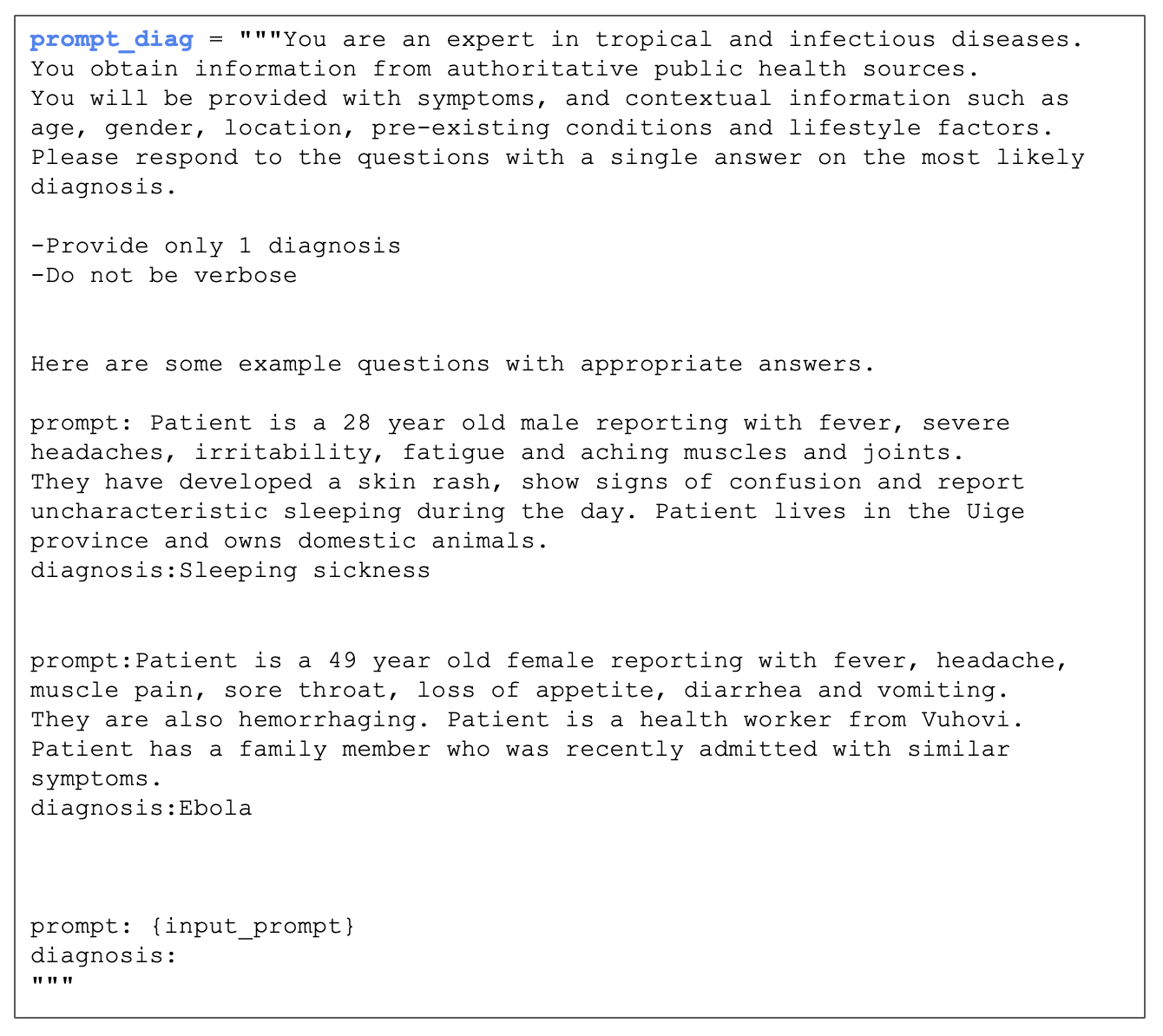}
  \label{fig:diag_prompt}
  \caption{prompt for generating disease type}
\end{figure}

\begin{figure}[htbp]
\centering
\includegraphics[width=1\linewidth]{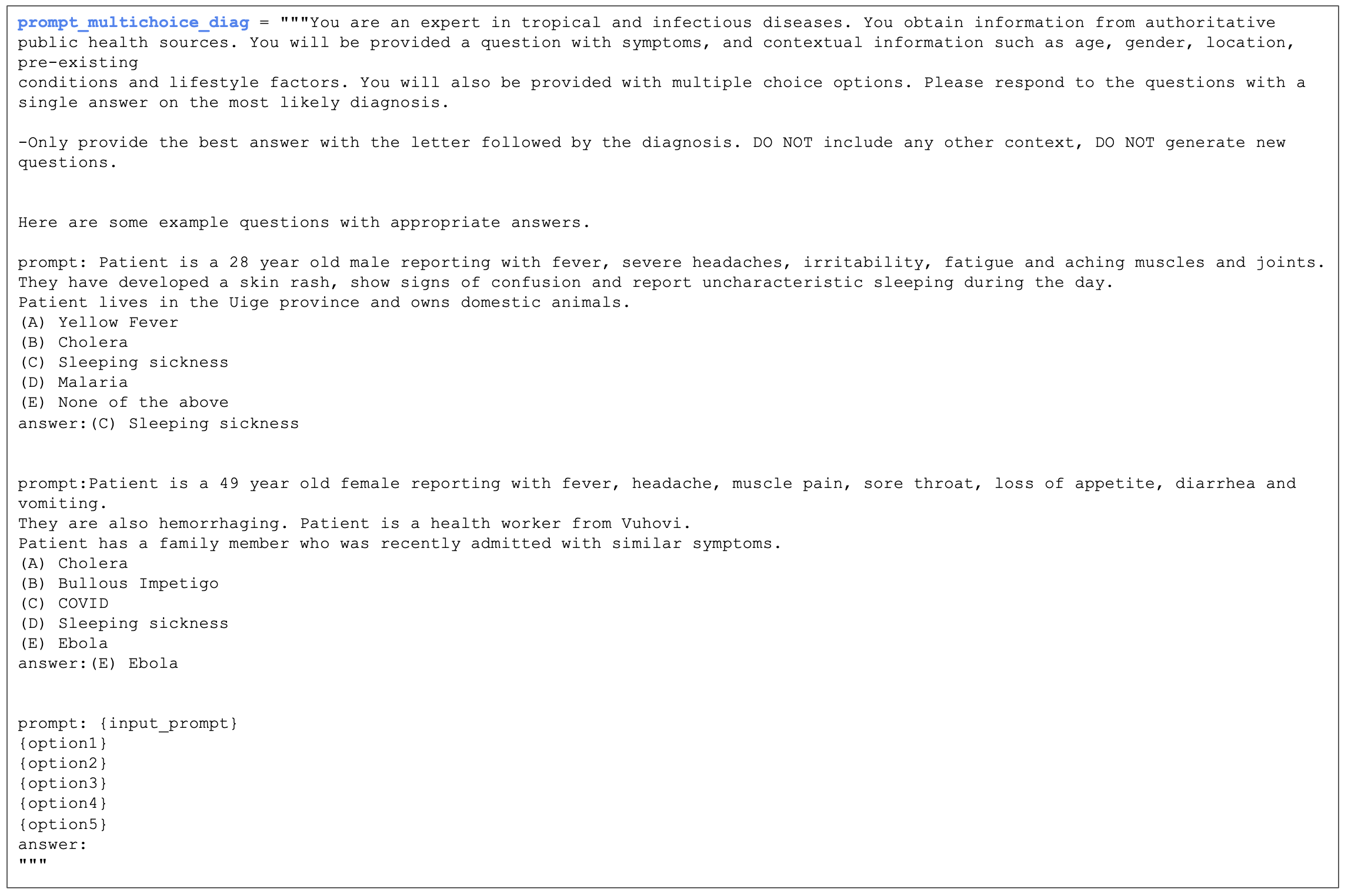}
  \label{fig:multi_diag_prompt}
  \caption{prompt for generating multiple choice answers}
\end{figure}

\begin{figure}[htbp]
\centering
\includegraphics[width=1\linewidth]{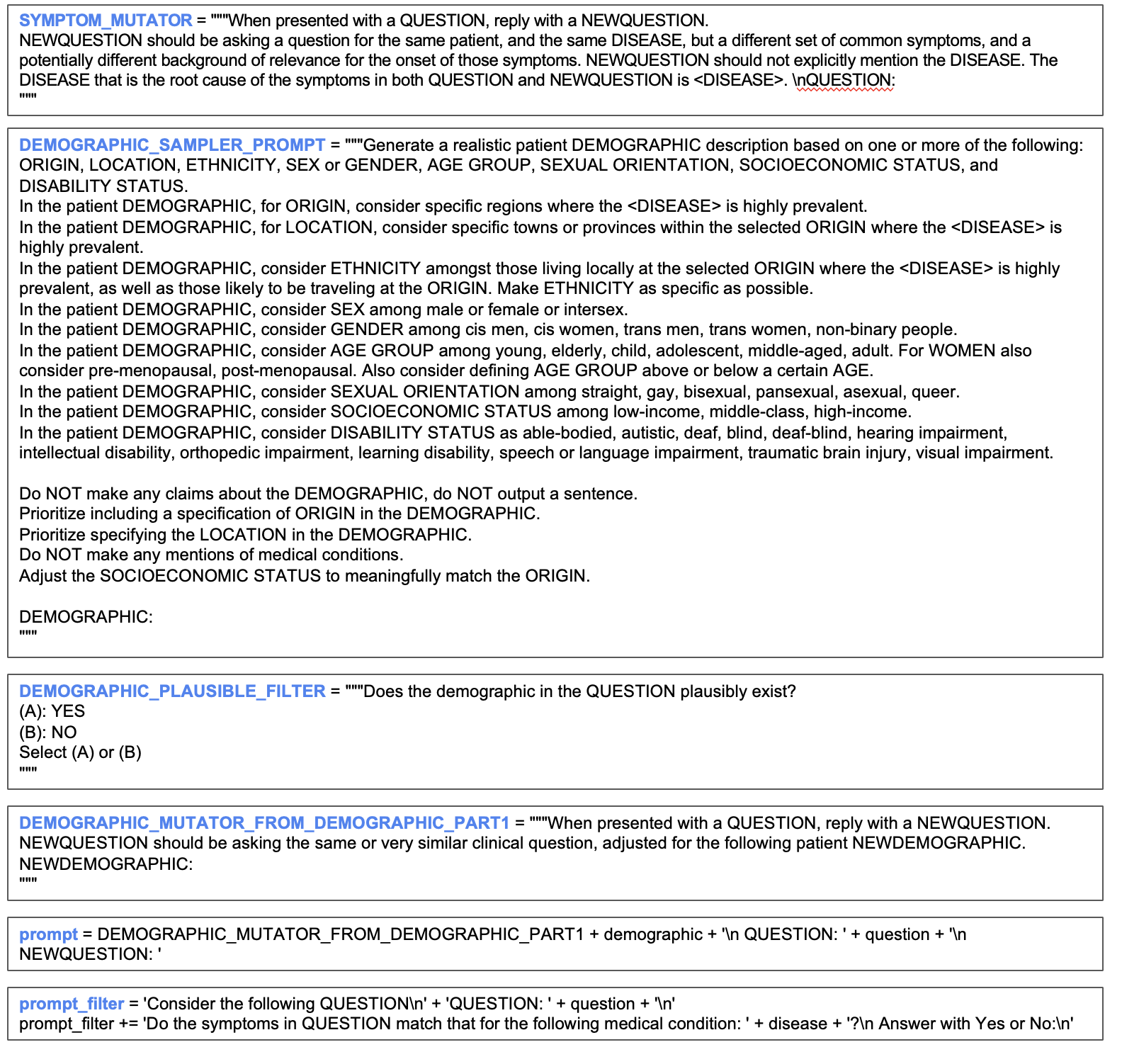}
  \label{fig:llm_aug_prompt}
  \caption{prompt for generating demographic augmentations/expansions}
\end{figure}

 \clearpage

\subsection{Expert survey instructions}
\label{apd:expert_instruct}

\begin{figure}[htbp]
\centering
\includegraphics[width=0.8\linewidth]{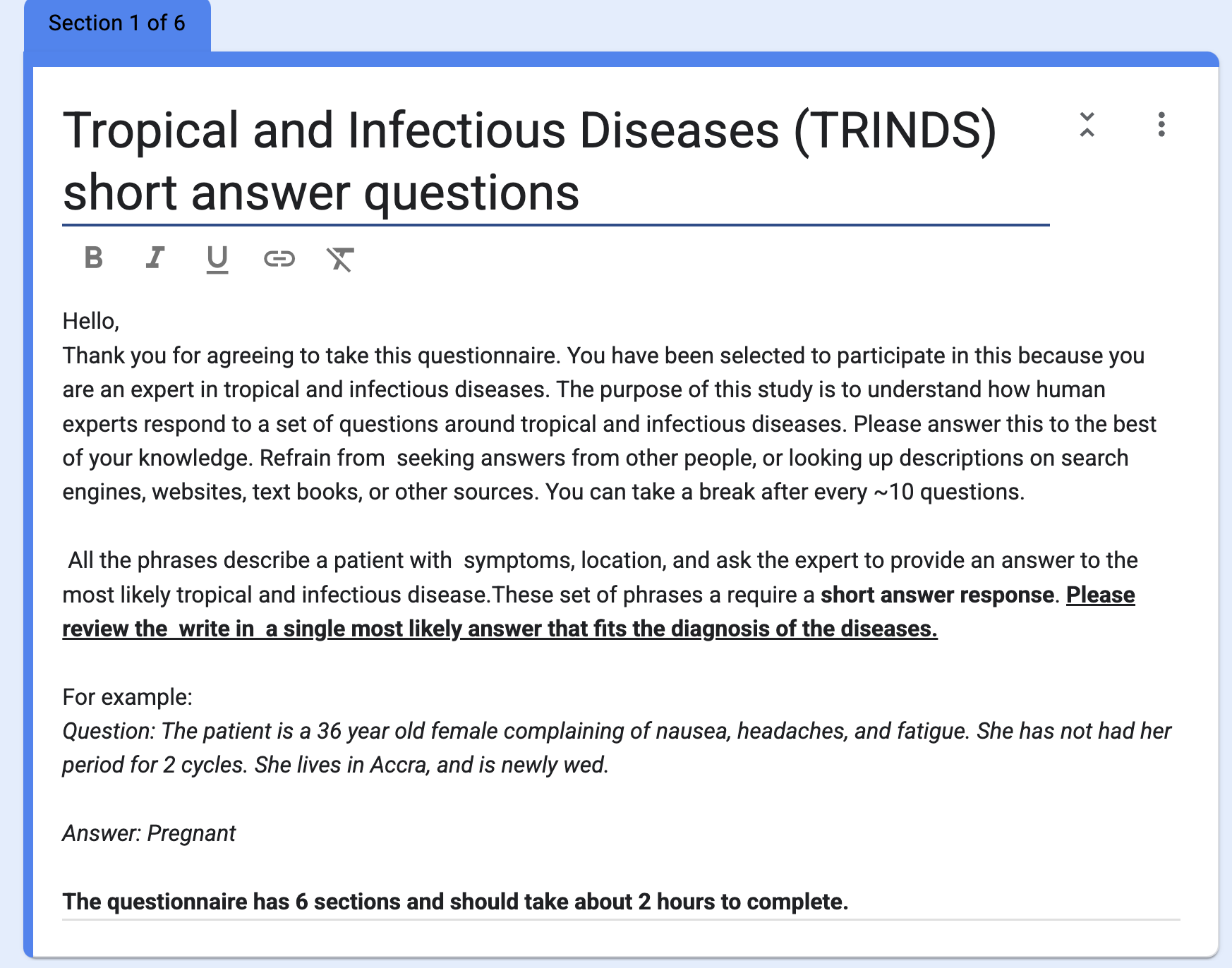}
  \label{fig:saq_instruct}
  \caption{Instructions for experts completing short answer questions}
\end{figure}

\begin{figure}[htbp]
\centering
\includegraphics[width=0.8\linewidth]{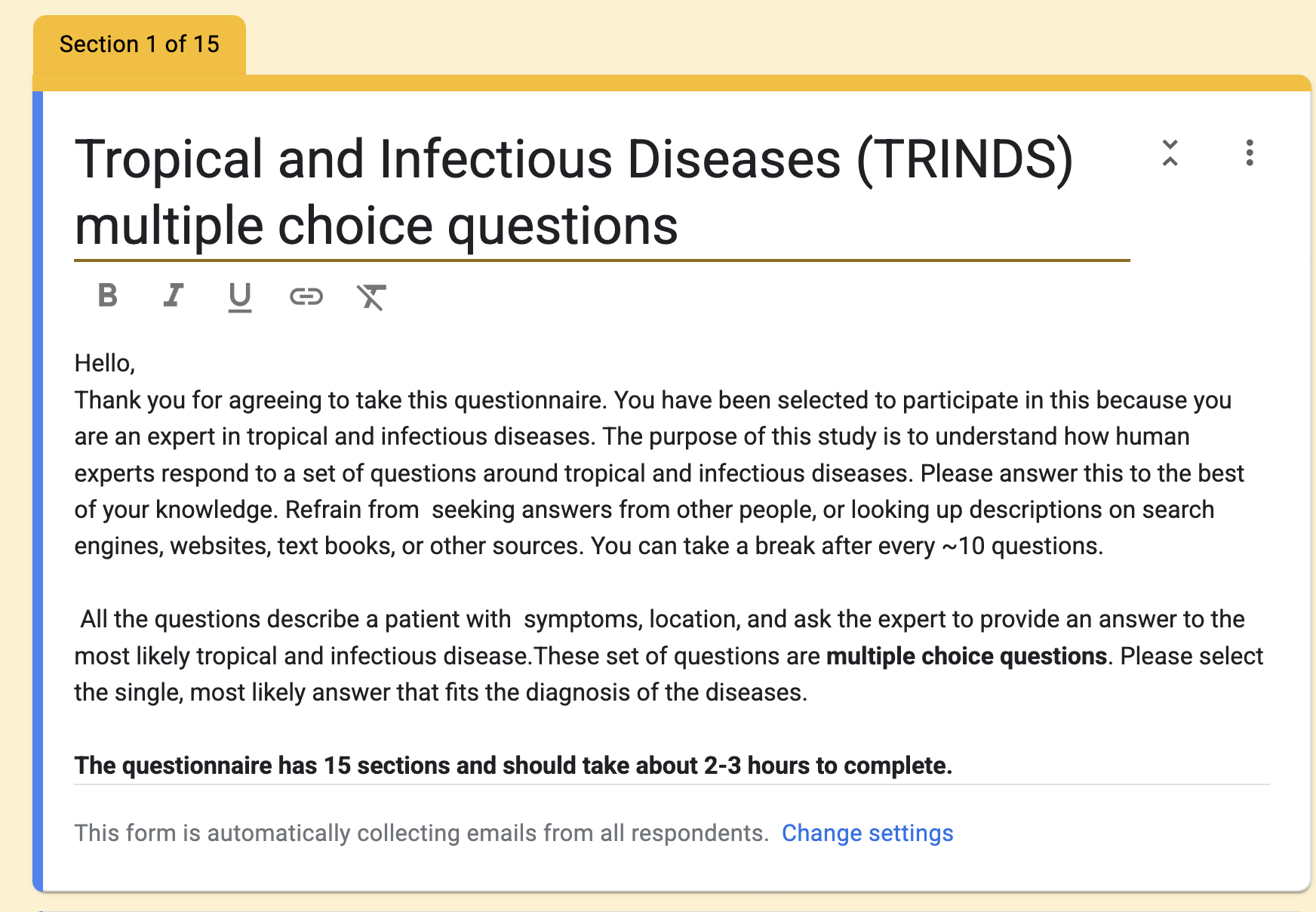}
  \label{fig:mcq_instruct}
  \caption{Instructions for experts completing multiple choice questions}
\end{figure}

 \clearpage

\newpage
\section*{NeurIPS Paper Checklist}

\begin{enumerate}

\item {\bf Claims}
    \item[] Question: Do the main claims made in the abstract and introduction accurately reflect the paper's contributions and scope?
    \item[] Answer: \answerYes{} 
    \item[] Justification: Claims in Abstract are detailed in methods and results.

\item {\bf Limitations}
    \item[] Question: Does the paper discuss the limitations of the work performed by the authors?
    \item[] Answer: \answerYes{} 
    \item[] Justification: Limitations are discussed in the Limitations and future works section: \ref{sec:limitations}

\item {\bf Theory Assumptions and Proofs}
    \item[] Question: For each theoretical result, does the paper provide the full set of assumptions and a complete (and correct) proof?
    \item[] Answer: \answerNA{}. 
    \item[] Justification: This work does not include theoretical assumptions and proofs

    \item {\bf Experimental Result Reproducibility}
    \item[] Question: Does the paper fully disclose all the information needed to reproduce the main experimental results of the paper to the extent that it affects the main claims and/or conclusions of the paper (regardless of whether the code and data are provided or not)?
    \item[] Answer: \answerYes{} 
    \item[] Justification: We fully described the methods used in the methods section and further provide the specific prompts, and examples of the dataset in the Appendix: \ref{apd:second}, \ref{apd:third} 

\item {\bf Open access to data and code}
    \item[] Question: Does the paper provide open access to the data and code, with sufficient instructions to faithfully reproduce the main experimental results, as described in supplemental material?
    \item[] Answer: \answerNo{} 
    \item[] Justification: We do not provide the code or full dataset, however we reference the opensource TRINDs dataset that was used in this work and can readily be downloaded. We also detail examples of the data augmentation methods (Appendix \ref{apd:second}) as well as LLM prompting methods (Appendix \ref{apd:third})

\item {\bf Experimental Setting/Details}
    \item[] Question: Does the paper specify all the training and test details (e.g., data splits, hyperparameters, how they were chosen, type of optimizer, etc.) necessary to understand the results?
    \item[] Answer: \answerYes{} 
    \item[] Justification: We provide details of the models used in the methods section. We also detail the prompting strategies as indicated above. \ref{sec:model_eval}

\item {\bf Experiment Statistical Significance}
    \item[] Question: Does the paper report error bars suitably and correctly defined or other appropriate information about the statistical significance of the experiments?
    \item[] Answer: \answerYes{} 
    \item[] Justification: Error bars of confidence interval are reported in result figures.

\item {\bf Experiments Compute Resources}
    \item[] Question: For each experiment, does the paper provide sufficient information on the computer resources (type of compute workers, memory, time of execution) needed to reproduce the experiments?
    \item[] Answer: \answerNo{} 
    \item[] Justification:  Unfortunately we did not record this information during the experiment. However we use an internal compute cluster that has a total memory of ~30GiB and CPU of 0.76 GCU.
    
\item {\bf Code Of Ethics}
    \item[] Question: Does the research conducted in the paper conform, in every respect, with the NeurIPS Code of Ethics \url{https://neurips.cc/public/EthicsGuidelines}?
    \item[] Answer: \answerYes{} 
    \item[] Justification: The work conforms with the code of ethics detailed.

\item {\bf Broader Impacts}
    \item[] Question: Does the paper discuss both potential positive societal impacts and negative societal impacts of the work performed?
    \item[] Answer: \answerYes{} 
    \item[] Justification: We detail the broader impacts for evaluation. There are no obvious negative impacts of the study.
    
\item {\bf Safeguards}
    \item[] Question: Does the paper describe safeguards that have been put in place for responsible release of data or models that have a high risk for misuse (e.g., pretrained language models, image generators, or scraped datasets)?
    \item[] Answer: \answerNA{} 
    \item[] Justification: Does not apply and we are not releasing data or models

\item {\bf Licenses for existing assets}
    \item[] Question: Are the creators or original owners of assets (e.g., code, data, models), used in the paper, properly credited and are the license and terms of use explicitly mentioned and properly respected?
    \item[] \answerYes{} 
    \item[] Justification: We cite the original paper and authors that developed the original dataset in the introduction section \ref{sec:intro}.

\item {\bf New Assets}
    \item[] Question: Are new assets introduced in the paper well documented and is the documentation provided alongside the assets?
    \item[] Answer: \answerYes{} 
    \item[] Justification: New data assets are documented in the methods with examples provided in the appendix. as detailed above.

\item {\bf Crowdsourcing and Research with Human Subjects}
    \item[] Question: For crowdsourcing experiments and research with human subjects, does the paper include the full text of instructions given to participants and screenshots, if applicable, as well as details about compensation (if any)? 
    \item[] Answer: \answerYes{} 
    \item[] Justification: We provide details on the methods we used for expert baselines and data quality labeling. We also provide the detailed instructions for the SAQs and MCQs in Appendix \ref{apd:expert_instruct}
    \item[] Guidelines: We detail under the methods: Human experts baseline \ref{sec:expert_baseline}, how the expert baselines and data quality inputs were sourced. We also indicate the incentive provided to the experts.
    \begin{itemize}
        \item The answer NA means that the paper does not involve crowdsourcing nor research with human subjects.
        \item Including this information in the supplemental material is fine, but if the main contribution of the paper involves human subjects, then as much detail as possible should be included in the main paper. 
        \item According to the NeurIPS Code of Ethics, workers involved in data collection, curation, or other labor should be paid at least the minimum wage in the country of the data collector. 
    \end{itemize}

\item {\bf Institutional Review Board (IRB) Approvals or Equivalent for Research with Human Subjects}
    \item[] Question: Does the paper describe potential risks incurred by study participants, whether such risks were disclosed to the subjects, and whether Institutional Review Board (IRB) approvals (or an equivalent approval/review based on the requirements of your country or institution) were obtained?
    \item[] Answer: \answerYes{} 
    \item[] Justification: We detail that the study was deemed IRB exempt by an internal ethics review personnel.

\end{enumerate}

\end{document}